\documentclass{CVM} 

\CVMsetup{
type      = {Research/Review Article},
doi       = {CVM.XXXX},
title     = {CORE-Seg: Reasoning-Driven Segmentation for Complex Lesions via Reinforcement Learning},
author    = {Yuxin Xie$^{1}$, Yuming Chen$^{1}$, Yishan Yang$^{1}$, Yi Zhou$^{1}$ \cor{}, Tao Zhou$^{2}$, Zhen Zhao$^{3}$, Jiacheng Liu$^{3}$, Huazhu Fu$^{4}$\\
},
runauthor = {Y. Xie, Y. Chen, Y. Yang, et al.},
abstract  = {
  Medical image segmentation is shifting from visual pattern matching toward cognitive reasoning. Although Multimodal Large Language Models (MLLMs) have shown promise in integrating linguistic and visual knowledge, significant gaps remain: existing general MLLMs possess broad common sense but lack the specialized visual reasoning required for complex lesions, whereas traditional segmentation models excel at pixel-level segmentation but lack logical interpretability. In this paper, we introduce ComLesion-14K, the first diverse Chain-of-Thought benchmark for reasoning-driven complex lesion segmentation. To accomplish this task, we propose CORE-Seg, an end-to-end framework integrating reasoning with segmentation through a Semantic-Guided Prompt Adapter. We design a progressive training strategy from Supervised Fine-Tuning to Group Relative Policy Optimization, equipped with an adaptive dual-granularity reward mechanism. CORE-Seg achieves state-of-the-art performance among MLLM frameworks, reaching 37.06\% mean Dice and an 18.42\% failure rate.  Project Page: \href{https://xyx1024.github.io/CORE-Seg.github.io/}{https://xyx1024.github.io/CORE-Seg.github.io/}.
},
keywords  = {Lesion Segmentation, Multimodal Reasoning, Multimodal Foundation Models, Reinforcement Learning},
copyright = {The Author(s) 2026.},
}

\begin{document}

\maketitle

\enlargethispage{-3pt}
\begin{figure}[b] \vskip -4mm
\small\renewcommand\arraystretch{1.3}
\begin{tabular}{p{80.5mm}} \toprule\\ \end{tabular}
\vskip -4.5mm \noindent \setlength{\tabcolsep}{1pt}
\begin{tabular}{p{3.5mm}p{80mm}}
$1\quad $ & School of Computer Science and Engineering, Southeast University, Nanjing, 211189, China. E-mail: silver\_iris@163.com; 220252314@seu.edu.cn;  ysyang0327@seu.edu.cn; yizhou.szcn@gmail.com \cor{}.\\
$2\quad $ & School of Computer Science and Engineering, Nanjing University of Science and Technology, Nanjing, 210094, China. E-mail: taozhou.ai@gmail.com.\\
$3\quad $ & Zhongda Hospital, Southeast University, Nanjing, 210009, China. E-mail: zhaozhen8810@126.com; jiachengliu@seu.edu.cn.\\
$4\quad $ & Institute of High-Performance Computing, Agency for Science, Technology and Research, 138632, Singapore. E-mail: hzfu@ieee.org.\\
&\hspace{-5mm} Manuscript received: 2025-01-01; accepted: 2025-01-01\vspace{-2mm}
\end{tabular} \vspace {-3mm}
\end{figure}

\section{Introduction}
Medical image segmentation~\cite{xie2024simtxtseg,rahman2024multi,wang2023weakly} is a core task of clinical diagnosis and treatment planning. However, conventional methods are confined to extracting visual patterns from input images, unable to model the cognitive reasoning process of clinicians. Thus, they often achieve limited performance on complex and ambiguous lesions requiring professional expertise. Recently, Multimodal Large Models (MLLMs)~\cite{li2023llava,jiang2025hulu} have offered a new research perspective by combining visual perception with linguistic understanding, showing potential for constructing more reasonable segmentation solutions.

\begin{figure}[!t]
\centerline{\includegraphics[width=\columnwidth]{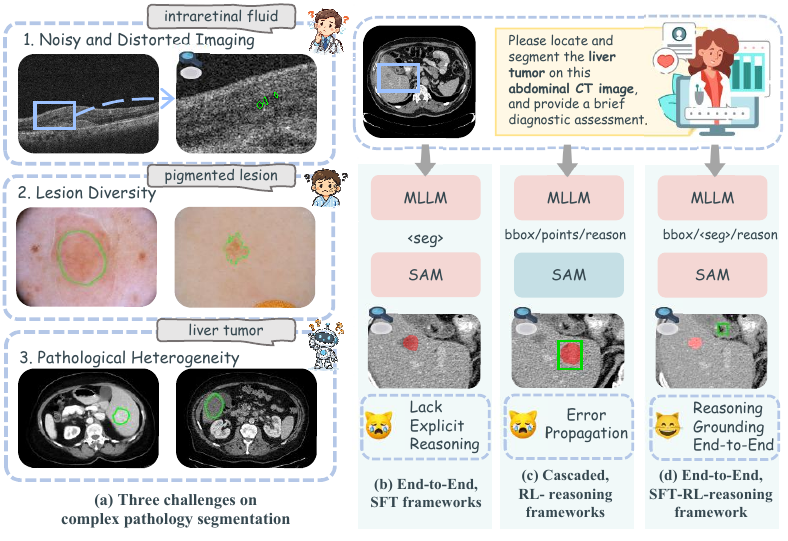}}
\caption{(a) Three major challenges in complex medical lesion segmentation, containing noisy and distorted imaging, lesion diversity, and pathological heterogeneity. (b) Supervised fine-tuned, end-to-end dense prediction architecture. (c) Reinforcement learning-driven, cascaded reasoning framework. (d) Our proposed end-to-end complex-lesion-centric reasoning segmentation  framework, CORE-Seg.}
\label{fig:intro}
\end{figure}

Despite the promising potential of MLLMs, current research~\cite{huang2025medseg,zou2025uncertainty, shen2026r2} mainly focus on commonsense reasoning and organ segmentation, while paying insufficient attention to reasoning-based segmentation for complex lesions. For instance, tasks like ``find the organ damaged by smoking", rely mostly on commonsense knowledge rather than on reasoning from specific visual cues, which limits their clinical value. Moreover, many studies focus on anatomical organs segmentation, which typically exhibit fixed locations and clear boundaries, presenting relatively low difficulty. In contrast, complex lesion regions that truly necessitate reasoning and are vital for diagnosis, have received less attention. Accordingly, research attention needs to be shifted from \textit{commonsense-driven organ segmentation} to \textit{visual reasoning oriented complex lesion segmentation}.

In this paper, we address this gap by focusing on \textbf{Complex Lesion Segmentation}, where intricate visual cues require explicit reasoning. As illustrated in Fig.~\ref{fig:intro} (a), complex lesions differ from the anatomical structures and apparent lesions commonly studied in existing methods, and pose three major challenges:
(1) \textit{Noisy and Distorted Imaging:} Degraded visual quality caused by acquisition noise and artifacts undermines stable feature extraction.
(2) \textit{Lesion Diversity:} Significant variations in lesion shape, location, and boundary limit the generalization of pattern-matching approaches.
(3) \textit{Pathological Heterogeneity:} Poor contrast and blurred lesion boundaries obscure the separation between targets and background.
These characteristics introduce substantial visual uncertainty, making it difficult for conventional pattern-matching models to achieve robust perception. To address these challenges, we introduce \textbf{ComLesion-14K}, the first large-scale Chain-of-Thought (CoT) benchmark designed for complex lesion segmentation. ComLesion-14K contains 14k cases covering 29 diseases, selected from 300k images to focus on challenging cases where conventional models often fail.

With this challenging benchmark in place, the limitations of existing solutions become even more apparent. Existing MLLM-based solutions generally follow two technical paths, yet both face significant hurdles in complex lesion segmentation. The first is the supervised fine-tuned (SFT), end-to-end dense prediction architecture (Fig.~\ref{fig:intro} (b), e.g., LISA~\cite{lai2024lisa}, MedPLIB~\cite{huang2025towards}). Although achieving a unified pipeline, its reasoning process is often implicit, obscuring crucial decision steps. Consequently, these models offer limited explainability and often struggle to segment ambiguous lesions that require deeper semantic understanding. The second is the reinforcement learning-driven, cascaded reasoning framework (Fig.~\ref{fig:intro} (c), e.g., Seg-Zero~\cite{liu2025seg}, MedReasoner~\cite{yan2025medreasoner}). It typically leverages the Group Relative Policy Optimization (GRPO) paradigm~\cite{shao2024deepseekmath} to enhance its reasoning capability by optimizing policies based on group-relative rewards. In this pipeline, the MLLM first generates a bounding box, which is then used as a prompt for a downstream segmenter such as SAM. However, this fragmented design suffers from severe error propagation—the final segmentation result is highly dependent on the initial localization accuracy, which is unreliable for complex clinical cases.

To overcome these limitations, we introduce \textbf{CORE-Seg, an end-to-end COmplex-lesion-centric REasoning Segmentation architecture} (Fig.~\ref{fig:intro} (d)). CORE-Seg combines the advantages of end-to-end modeling and explicit reasoning empowered by reinforcement learning. This design avoids the fragmented pipeline of cascaded methods and renders the reasoning process more interpretable than traditional supervised segmentation frameworks.  Specifically, we introduce a Semantic-Guided Prompt Adapter, which maps the hidden states of the special \texttt{<seg>} token from the textual space of the MLLM to the visual feature space of SAM. This allows the model to leverage semantic reasoning to directly guide mask prediction, eliminating the error propagation. We also develop a progressive training pipeline evolving from SFT to GRPO. An adaptive reward mechanism is further introduced to alleviate reward sparsity and jointly boost the capabilities of lesion reasoning, localization and mask prediction. With this design, CORE-Seg generates explainable reasoning while maintaining the robustness of end-to-end segmentation, achieving better performance on complex lesion segmentation tasks.

The key contributions of this work are fourfold:

(1) We define a new task paradigm—\textbf{Complex Lesion Segmentation}, which necessitates reasoning-driven understanding of lesion regions characterized by high visual ambiguity and heterogeneity.

(2) We construct \textbf{ComLesion-14K}, the first benchmark for complex lesion segmentation. It is curated from challenging clinical cases that existing methods struggle to localize. With CoT annotations, the dataset fills the research gap in interpretable complex medical image segmentation.

(3) We propose \textbf{CORE-Seg}, an end-to-end reasoning framework anchored by a \textbf{Semantic-Guided Prompt Adapter}. This module bridges MLLM reasoning with visual segmentation to eliminate cascading errors. Furthermore, we devise a \textbf{progressive SFT-to-GRPO pipeline} and an \textbf{adaptive dual-granularity reward mechanism}, effectively resolving reward sparsity while ensuring precise and interpretable predictions.

(4) CORE-Seg establishes a new state-of-the-art among MLLM frameworks for complex lesion segmentation. Extensive experiments demonstrate that our framework outperforms leading baselines by a significant margin (achieving 37.06\% mDice, a +14.89\% improvement), while reducing the failure rate to 18.42\%.

\section{Related Work}
\subsection{Multimodal Medical Segmentation Benchmarks}
With the rapid advancement of MLLMs, a multitude of reasoning methods and Visual Question Answering (VQA) datasets have swiftly emerged in both natural~\cite{hurst2024gpt,hu2025groundingsuite} and medical domains~\cite{zhang2024development,hu2024omnimedvqa,wu2025bridging, chen2026beyond}. In the field of medical image segmentation, large-scale datasets such as SA-Med2D-20M~\cite{ye2023sa} and IMed-361M~\cite{Cheng_2025_CVPR} serve as foundational visual corpora for instruction tuning. BiomedParseData~\cite{zhao2025foundation} features 6.2M image-mask-text triplets for referring segmentation, yet it is limited to explicit naming and lacks reasoning support. U-MRG-14K~\cite{yan2025medreasoner} is the first cross-modal medical dataset to integrate implicit clinical queries, CoT reasoning traces, and pixel-level masks across 14k samples. However, reasoning benchmarks are often confined to single-target organs, while referring datasets lack explanatory capabilities. This creates a gap in reasoning analysis of complex multi-focal lesions. To address this, we propose the ComLesion-14K dataset with CoT annotations for diverse lesion study and multi-target segmentation.

\begin{figure*}[!t]
\centerline{\includegraphics[width=\linewidth]{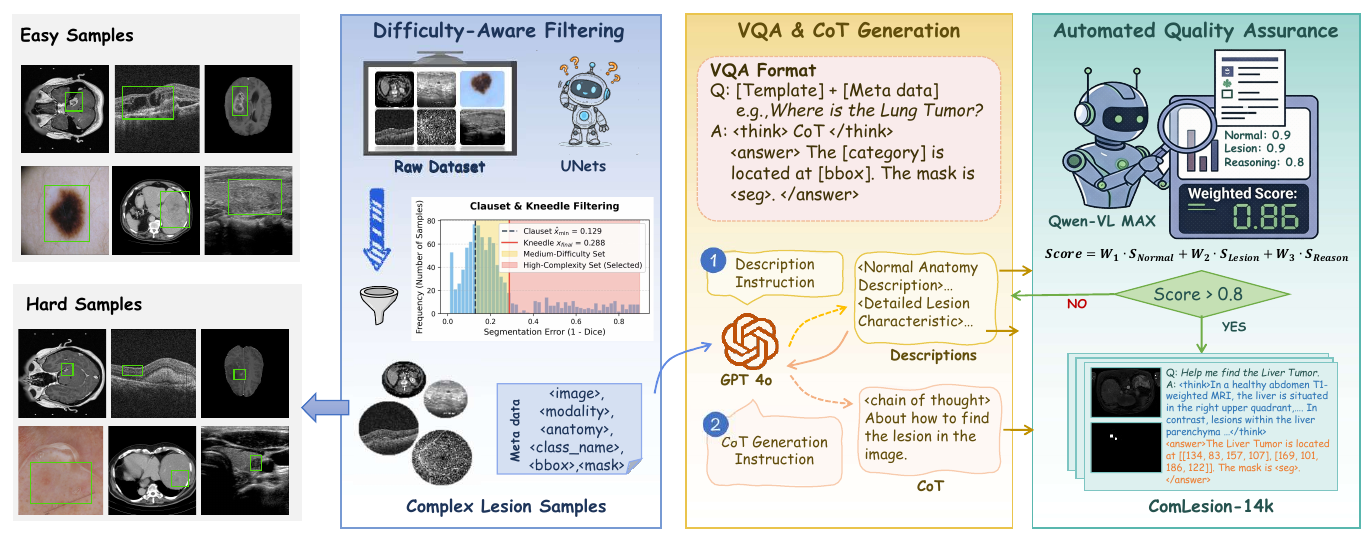}}
\caption{\textbf{Overview of the ComLesion-14K dataset construction pipeline.}
The pipeline includes difficulty-aware filtering, VQA \& CoT generation, and automated quality assurance.
Easy and hard samples are compared on the left to demonstrate the necessity of filtering mechanism.
Bounding boxes are adopted to clearly visualize lesion morphology without obscuring details.}
\label{fig:dataset_gen}
\end{figure*}

\subsection{Reasoning-Enhanced Visual Grounding}
In recent years, specialist medical segmentation systems have achieved strong performance, such as robust segmentation architectures~\cite{hu2026pranet, azad2025transception, gama2021learning}, as well as disease-oriented frameworks~\cite{dai2021deep,li2024integrated}. Although effective for specific clinical workflows, these methods are usually designed for fixed diseases and predefined label spaces, thus suffering from limited generalization and weak interpretability.

The advent of foundation models (e.g., CLIP~\cite{radford2021learning}, SAM~\cite{kirillov2023segment}) facilitated referring segmentation approaches including BiomedParse and CLIP-Driven Universal Model~\cite{liu2023clip}. While these methods utilize text for guidance, they function primarily as recognition modules, failing to provide rational explanations for their decisions.

The rise of Large Language Models has broken this bottleneck. Inspired by LISA, which endowed segmentation with reasoning capabilities, medical reasoning models have rapidly evolved. For instance, MedPLIB pioneered pixel-level reasoning via a multi-expert strategy, while MediSee~\cite{tong2025medisee} incorporated logic-driven constraints for superior performance. However, these methods often rely heavily on instruction tuning, making them prone to overfitting. Crucially, they lack the capability to generate explicit reasoning processes for complex scenarios—a limitation our work aims to address.

\subsection{Reinforcement Learning for Image Segmentation}
Reinforcement learning (RL) has become an important approach for aligning model behavior with human preferences. Its development has progressed from Proximal Policy Optimization (PPO)~\cite{schulman2017proximal} to Direct Preference Optimization (DPO)~\cite{rafailov2023direct}, while GRPO provides a more efficient optimization strategy. Recent medical reasoning models like Med-R1~\cite{lai2026med} and Colon-X~\cite{ji2025colon} further show the potential of RL in medical applications.

In medical image segmentation, however, existing RL-based methods such as Seg-Zero, SAM-R1~\cite{huang2025sam}, and MedReasoner mostly follow a serial pipeline. In these approaches, the reasoning component and the segmentation module are optimized separately, which can lead to error propagation and reduce the consistency between reasoning and prediction. Inspired by LENS~\cite{zhu2025lens}, we instead adopt a unified framework that combines SFT with GRPO, allowing explicit medical reasoning and pixel-level segmentation to be optimized jointly.

\section{ComLesion-14K Dataset}
\label{sec:dataset}
\subsection{Overview}
Most existing datasets for medical image segmentation typically focus on specific modalities with fixed labels. While recent studies have introduced visual prompts (e.g., points, boxes) or text guidance (e.g., U-MRG, MLMR-SD~\cite{tong2025medisee}), they mainly target organ-level structures, overlooking the pathological details required for lesion identification. Complex lesion segmentation is clinically critical yet more challenging due to subtle boundaries and intricate anatomical contexts.

To bridge this gap, we introduce \textbf{ComLesion-14K}, a multimodal dataset specifically designed for complex lesion segmentation and prompt-driven semantic segmentation. Unlike existing VQA segmentation datasets, it supports category-specific lesion segmentation, which segments all corresponding lesion instances according to given textual prompts. It also incorporates explicit reasoning chains that mimic the clinician's diagnosis process—identifying global context, localizing lesions, and delineating boundaries. The construction pipeline comprises three stages, as illustrated in Fig.~\ref{fig:dataset_gen}.

\subsection{Data Selection and Difficulty-Aware Filtering}
To curate challenging samples from 23 public datasets, we adopt a \textbf{difficulty-aware filtering strategy} rather than a static threshold. For each lesion class, three U-Net models~\cite{ronneberger2015u} with identical configurations and different random seeds are trained. The average Dice score of the three predictions is calculated for each sample to mitigate training randomness. The sample difficulty is defined as: $x = 1 - \text{Dice}$.

First, the \textbf{Clauset method}~\cite{clauset2009power} is employed to estimate the initial threshold $\hat{x}_{\min}$ for the long-tail subset. By fitting a power-law distribution to the upper-tail samples ($x_i \geq x_{\min}$), the optimal $\hat{x}_{\min}$ is determined by minimizing the Kolmogorov-Smirnov (KS) distance:
\begin{equation}
\hat{x}_{\min} = \arg\min_{x_{\min}} \left( \max_{x \geq x_{\min}} |S(x) - P(x)| \right),
\end{equation}
where $S(x)$ and $P(x)$ denote the empirical and fitted cumulative distributions for the tail, respectively.

We then derive the final filtering threshold via \textbf{an adaptive elbow detection algorithm~\cite{satopaa2011finding}}, expressed as:
\begin{equation}
x_{\text{final}} = x\left[ \arg\max_{i} \,\Big( y_{\text{norm}}(i)  - x_{\text{norm}}(i) \Big) \right]
\end{equation}
This algorithm robustly separates the challenging long-tail subset ($x \geq x_\text{final}$) from the moderate-difficulty background, followed by a final manual inspection to filter out annotation artifacts.

Moreover, manual inspection is conducted to exclude samples with annotation artifacts. As shown in Fig.~\ref{fig:dataset_gen}, simple lesions have clear boundaries and large areas, whereas hard lesions suffer from blurred boundaries, small sizes and imaging noise, bringing greater segmentation challenges.

\begin{figure*}[!t]
\centerline{\includegraphics[width=\linewidth]{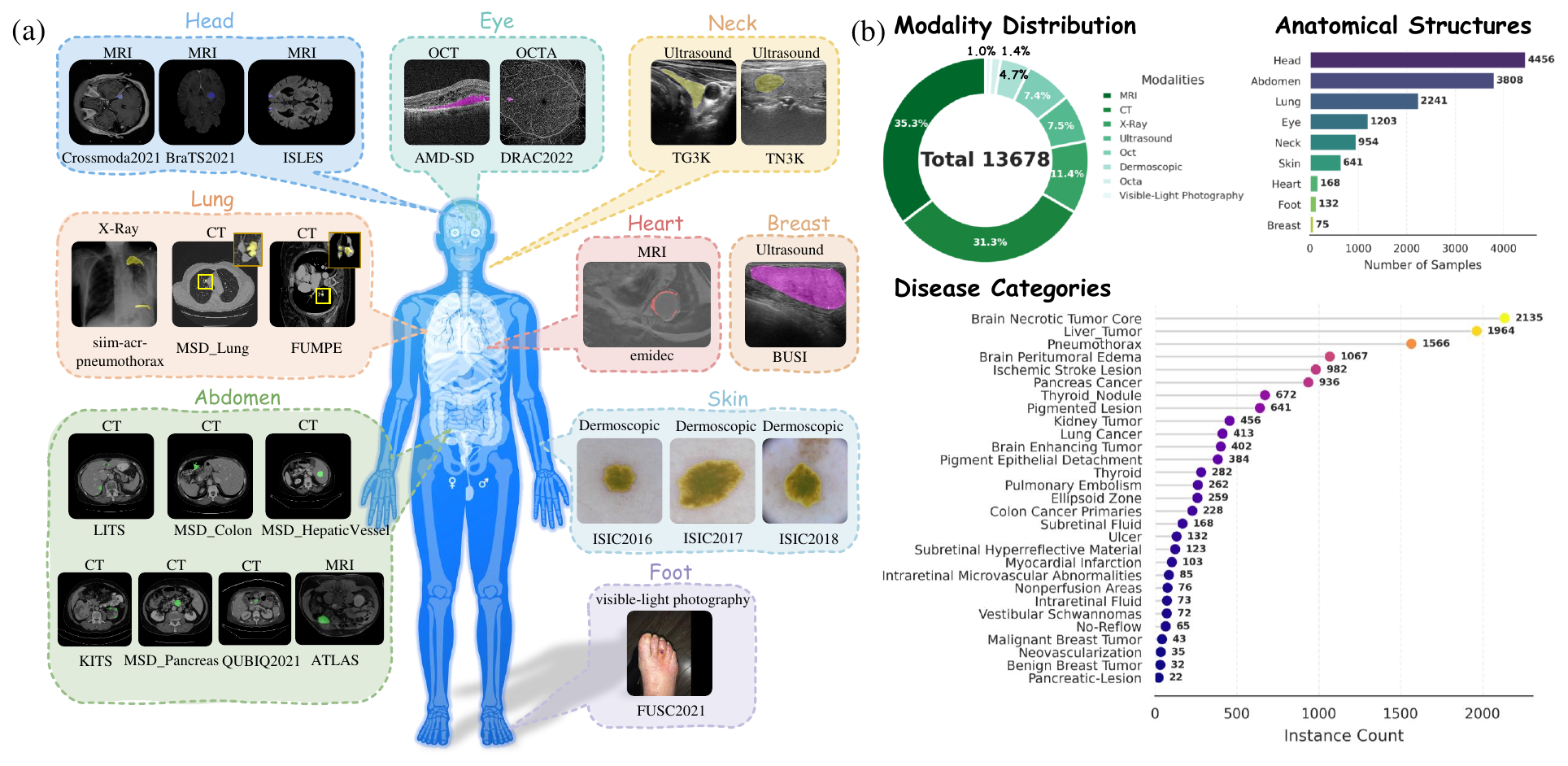}}
\caption{ (a) Representative samples illustrating broad organ coverage across diverse imaging modalities. (b) Dataset statistics summarizing the distribution of imaging modalities, anatomical structures, and disease categories.}
\label{fig:dataset}
\end{figure*}

\subsection{CoT and VQA Generation}
For each sample, we construct a structured tuple $\mathcal{D} = \{I, M, Q, R, A\}$. Here, $I$ and $M$ denote the raw image and segmentation mask. The textual components $Q,R,A$ are derived from metadata (modality, anatomy, disease category) and bounding boxes extracted from $M$. 

The clinical queries $Q$ are synthesized using \textbf{dynamic templates} to ensure linguistic diversity. The answers $A$ strictly follow the format: ``The \{lesion\} is located at \{bboxes\}.", serving as direct supervision for localization.

To generate high-quality reasoning text $R$, we leveraged GPT-4o~\cite{hurst2024gpt} with a two-step prompting strategy:
\begin{itemize}
    \item \textbf{Step 1: Initial Visual Observation.} Conditioned on the metadata, the model generates \textbf{a concise description} of the normal anatomical context and a \textbf{detailed characterization} of the lesion (e.g., specific location, irregular shape, heterogeneous texture).
    \item \textbf{Step 2: Visual Reasoning for Localization.} Given the target disease, this step focuses on the reasoning process of identification. The model leverages evidence from Step 1 to deduce the precise spatial location, explicitly explaining how to distinguish the target area from the surrounding background and confounders.
\end{itemize}
Crucially, explicit coordinate values are stripped from the final $R$. This constraint prevents the model from shortcut learning via numerical regression, forcing it to rely on semantic-level anatomical understanding.

\subsection{Automated Quality Assurance}
To guarantee the logical consistency of the generated text, we established an automated review pipeline using Qwen2.5-VL-Max. By decoupling the evaluator from the generator, this design avoids potential biases and style tendencies, ensuring that the evaluation results are more objective and reliable. The model scores each sample based on three criteria: normal anatomy description ($S_{normal}$), lesion characterization ($S_{lesion}$), and reasoning logic ($S_{reason}$). The final quality score $S_{final}$ is calculated as:
\begin{equation}
    S_{final} = w_1 \cdot S_{normal} + w_2 \cdot S_{lesion} + w_3 \cdot S_{reason},
\end{equation}
where we empirically set weights to $w_1=0.3, w_2=0.3, w_3=0.4$ to emphasize the importance of logic reasoning. Samples with $S_{final} < 0.8$ were regenerated.

\begin{table*}[t]
\centering
\caption{Detailed Composition of the ComLesion-14K Dataset}
\setlength{\tabcolsep}{2pt}
\label{tab:composition}
{%
\begin{tabular}{lcccp{3.5cm}c}
\toprule
Dataset Name & Samples & Modality & Anatomy & Annotation Type & Access \\
\midrule
BraTS2021~\cite{baid2021rsna} & 3604 & MRI & Head & Brain peritumoral edema, necrotic tumor core, enhancing tumor & \checkmark \\
SIIM-ACR~\cite{siim-acr-pneumothorax-segmentation} & 1566 & X-ray & Pleural cavity & Pneumothorax & \checkmark \\
MSD\_HepaticVessel~\cite{antonelli2022medical} & 1098 & CT & Abdomen & Liver Tumor & \checkmark \\
AMD-SD~\cite{hu2024amd} & 1007 & OCT & Eye & Pigment epithelial detachment, ellipsoid zone, intraretinal fluid, subretinal fluid & \checkmark \\
MSD\_Pancreas & 936 & CT & Abdomen & Pancreas Cancer & \checkmark \\
ISLES2022~\cite{hernandez2022isles} & 780 & MRI & Head & Ischemic Stroke Lesion & \checkmark \\
TN3K~\cite{gong2021multi} & 672 & Ultrasound & Neck & Thyroid Nodule & \checkmark \\
LITS~\cite{bilic2023liver} & 664 & CT & Abdomen & Liver Tumor & \checkmark \\
KiTS~\cite{heller2019kits19} & 658 & CT & Abdomen & Kidney Tumor, Kidney Cyst & \checkmark \\
MSD\_LungTumours & 413 & CT & Thoracic cavity & Lung cancer & \checkmark \\
ISIC2018~\cite{codella2019skin} & 389 & Dermoscopy & Skin & Pigmented lesion & \checkmark \\
TG3K~\cite{gong2023thyroid} & 282 & Ultrasound & Neck & Thyroid & \checkmark \\
FUMPE2018~\cite{masoudi2018new} & 262 & CT & Lung & Pulmonary Embolism & \checkmark\\
MSD\_Colon & 228 & CT & Abdomen & Colon cancer primaries & \checkmark\\
ISIC2017~\cite{codella2018skin} & 220 & Dermoscopy & Skin & Pigmented lesion & \checkmark \\
ATLAS~\cite{quinton2023tumour} & 202 & MRI & Abdomen & Liver tumor & \checkmark \\
DRAC2022~\cite{qian2024drac} & 196 & OCTA & Eye & Nonperfusion areas, intraretinal microvascular abnormalities, neovascularization & \checkmark \\
EMIDEC~\cite{lalande2020emidec} & 168 & MRI & Heart & Myocardial infarction, no-reflow area & \checkmark \\
FUSC~\cite{wang2024fuseg} & 132 & \makecell{Visible-light\\ photography} & Foot & Ulcer & \checkmark \\
BUSI~\cite{al2020dataset} & 75 & Ultrasound & Breast & Benign breast tumor, malignant breast tumor & \checkmark \\
CrossMoDa2021~\cite{shapey2021segmentation} & 72 & MRI & Head & Vestibular schwannomas & \checkmark \\
ISIC2016~\cite{gutman2016skin} & 32 & Dermoscopy & Skin & Pigmented lesion & \checkmark \\
QUBIQ2021~\cite{vzukovec2021modeling} & 22 & CT & Abdomen & Pancreatic lesion & \checkmark \\
\midrule
\textbf{Total} & \textbf{13678} & \textbf{8} & \textbf{9} & \textbf{29} & \textbf{23}\\
\bottomrule
\end{tabular}
}
\end{table*}

\subsection{Dataset Statistics and Distribution}
The ComLesion-14K dataset comprises 13678 samples that span 8 imaging modalities, 9 anatomical regions, and 29 disease categories. Its composition is shown in Table~\ref{tab:composition}. As shown in Fig.~\ref{fig:dataset}, the distribution reflects the availability of clinical data in real-world, with MRI and CT being the dominant modalities (accounting for $\sim$66\% of data), and head and abdomen being the primary anatomical regions. Despite this imbalance, the dataset covers a wide spectrum of lesion types, ranging from large organ tumors to subtle retinal lesions, ensuring the model's robustness across varied scales and contrasts.

\begin{figure*}[!t]
\centerline{\includegraphics[width=\linewidth]{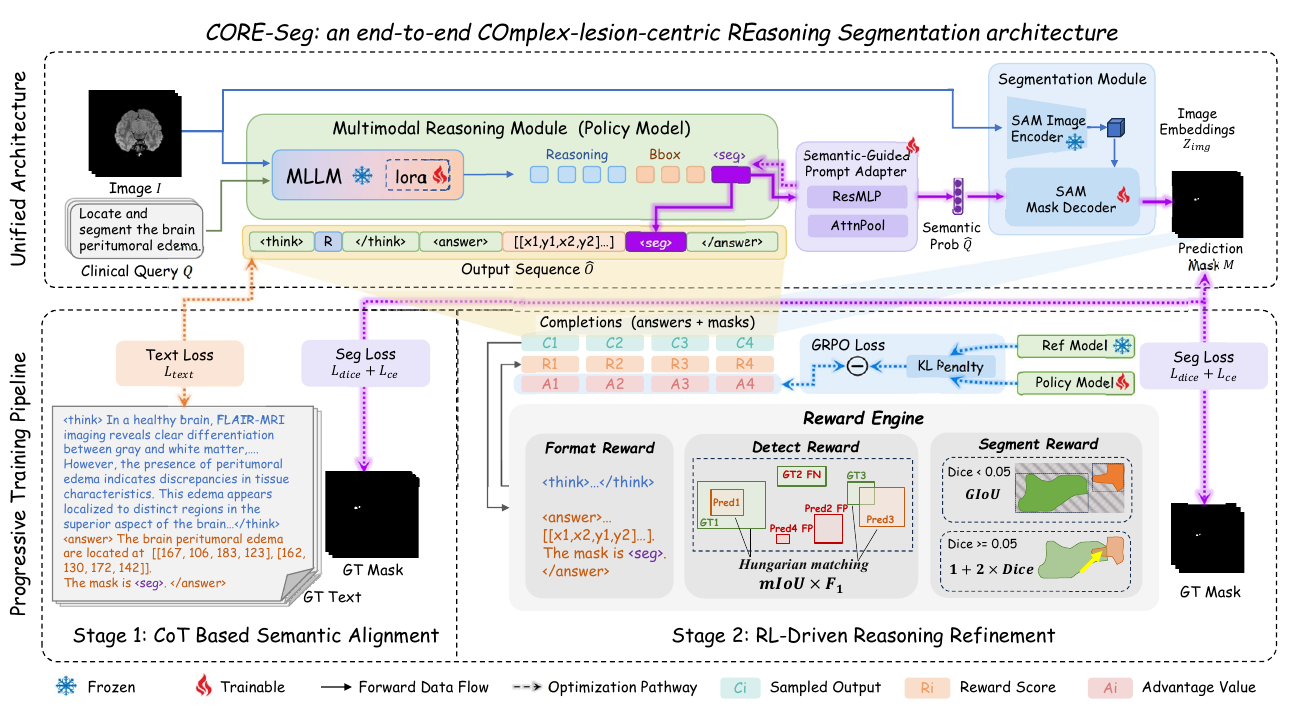}}
\caption{\textbf{Overview of the CORE-Seg framework.}
The architecture integrates an MLLM with a SAM via a Semantic-Guided Prompt Adapter. The training consists of two stages:
(1) CoT-Based Semantic Alignment Stage: Supervised fine-tuning establishes the mapping between clinical reasoning and lesion localization using the \texttt{<seg>} token.
(2) RL-Driven Reasoning Refinement Stage: The model is further optimized using GRPO. A multi-dimensional Reward Engine (evaluating format, detection, and segmentation) serves as the critic, where a Dual-Granularity Mask Reward penalizes disconnected regions and encourages precise boundary delineation.}
\label{fig:framework}
\end{figure*}

\section{CORE-Seg}
\subsection{Unified End-to-End Architecture}
Building upon ComLesion-14K, we propose CORE-Seg, an end-to-end multimodal framework that couples complex lesion segmentation with interpretable reasoning in a single fully differentiable network, thereby avoiding the error propagation of conventional cascaded pipelines such as Seg-Zero, MedReasoner, and SAM-R1 (Fig.~\ref{fig:framework}). It consists of three components: the Multimodal Reasoning Module, which interprets the medical context and localizes candidate lesions; the \textbf{Semantic-Guided Prompt Adapter}, which translates these semantic cues into a visual prompt; and the Segmentation Module, which converts that prompt into a pixel-level mask. 

The framework is trained in two stages: a CoT cold-start stage that aligns cross-modal semantics, and a reinforcement learning stage driven by an adaptive dual-granularity reward mechanism, which unlocks the reasoning potential of the model and refines segmentation precision.

\subsection{Multimodal Reasoning Module}
We employ an MLLM, specifically Qwen2.5-VL-3B, as the reasoning backbone. Given an input image $I$ and a clinical query $Q$, the model generates a structured output sequence $\hat{O}$ formatted as:

\begin{equation}
\begin{split}
\hat{O} &= \texttt{<think>} \,\Vert\, \hat{R} \,\Vert\, \texttt{</think>} \\
&\quad \,\Vert\, \texttt{</answer>} \,\Vert\, \hat{A} \,\Vert\, \texttt{</answer>}.
\end{split}
\end{equation}

Here, $\Vert$ denotes token sequence concatenation, $\hat{R}$ and $\hat{A}$ are reasoning and answer correspond strictly to the ground-truth $R$ and $A$ defined in Sec.~\ref{sec:dataset}.  $\hat{A}$ contains the special learnable token \texttt{<seg>} at its end to trigger the subsequent mask decoder.

The hidden state of this \texttt{<seg>} token acts as a semantic anchor, aggregating context from both reasoning rationales and localization cues via Transformer attention. By encapsulating ``what to segment" and ``where it is", the token provides a semantic-rich query for the Semantic-Guided Prompt Adapter to effectively retrieve pixel-level visual features.

\subsection{Semantic-Guided Prompt Adapter}
The Semantic-Guided Prompt Adapter bridges the high-level reasoning of the MLLM with pixel-level segmentation. Specifically, we extract the final-layer hidden state of the special \texttt{<seg>} token as a condensed semantic anchor ${P} \in \mathbb{R}^{B \times 1 \times D}$, where $B$ denotes the batch size and $D$ is the hidden dimension of the MLLM. This embedding summarizes the multimodal reasoning outcome and encodes implicit spatial priors. Unlike explicit coordinate prompts (e.g., points or boxes), As a global semantic descriptor, $P$ directs the decoder to attend to semantically consistent regions, allowing \textbf{multi-instance segmentation} for all spatially disjoint lesions of the given class.

To reduce the inherent feature distribution gap between language and vision, ${P}$ is first projected into a visually aligned manifold via stacked Residual MLP (ResMLP) blocks:
\begin{equation}
{h}_{l+1} = {h}_l + \mathrm{ResMLP}\left( \mathrm{LayerNorm}({h}_l) \right),
\end{equation}
where ${h}_0 = {P}$, and the final output denotes the refined feature $\tilde{{P}}$. To adapt this linguistic feature for compatibility with the segmentation decoder, we employ an attention module that utilizes a learnable query ${q}_{learn}$. This query extracts and reconstructs the segmentation prompt by engaging with $\tilde{{P}}$ through mechanisms of cross attention:
\begin{equation}
\hat{P} = \mathrm{Proj}\left( \mathrm{LayerNorm}\left( \mathrm{Attn}({q}_{learn}, \tilde{{P}}, \tilde{{P}}) \right) \right),
\end{equation}
where ${q}_{learn}$ serves as the query while $\tilde{{P}}$ acts as both key and value. The resulting semantic probe $\hat{P} \in \mathbb{R}^{B \times 1 \times 256}$ is then injected into the segmentation decoder, effectively modulating the mask generation process with precise linguistic guidance.

\subsection{Segmentation Model}
We use the pre-trained MedSAM2~\cite{ma2025medsam2} as the segmentation backbone. For an input medical image, the SAM image encoder (e.g., ViT-H) produces image embeddings $Z_{img}$. We then inject the semantic probe $\hat{P}$ into SAM’s prompt pathway. The mask decoder takes $(Z_{img}, \hat{P})$ as inputs and leverages SAM’s native two-way attention to fuse prompt and image features, yielding the binary segmentation mask $\hat{M} \in \mathbb{R}^{H \times W}$:
\begin{equation}
\hat{M} = \mathrm{SAM}_{\text{Decoder}}(Z_{img}, \hat{P}).
\end{equation}

Although SAM can be conditioned by the semantic probe, reliable semantic prompting requires both medical reasoning and robustness. To this end, we propose a two-stage training strategy that first establishes stable semantic alignment and further enhances generalization and segmentation performance.
 
\subsection{Stage 1: CoT-Based Semantic Alignment}
In this cold-start stage, we aim to inject domain-specific medical knowledge and establish a preliminary alignment between linguistic reasoning and visual localization. To achieve this efficiently, we apply Low-Rank Adaptation (LoRA) to the MLLM, enabling it to learn specialized CoT patterns while keeping most parameters frozen. Concurrently, we fine-tune the Semantic-Guided Prompt Adapter and the SAM mask decoder to ensure the semantic \texttt{<seg>} token is accurately translated into spatial queries for segmentation.

The training is supervised end-to-end using a composite loss function:
\begin{equation}
\mathcal{L}_{\text{S1}}
= \lambda_1 \mathcal{L}_{\text{text}}(\hat{O}, O)
+ \lambda_2 \mathcal{L}_{\text{DiceCE}}(\hat{M}, M).
\end{equation}

Here, $\mathcal{L}_{\text{text}}$ is the auto-regressive cross-entropy loss over the predicted output sequence $\hat{O}$, whose target sequence $O$ is formed from the ground-truth reasoning $R$ and answer $A$ with special tokens. $\mathcal{L}_{\text{DiceCE}}$ combines Dice loss and cross-entropy loss for segmentation. $ \{\lambda_i\}_{i=1}^2 $ are balancing coefficients. This joint objective encourages the model to capture medical semantics while enhancing its ability to localize lesion regions.

\subsection{Stage 2: RL-Based Reasoning Exploration and Segmentation Refinement}
\subsubsection{Reward Function}
Following the supervised alignment, we introduce RL to enable reasoning exploration and error correction on complex medical cases, while simultaneously refining segmentation boundaries. We employ GRPO as our core algorithm. Compared to the PPO, GRPO is more memory-efficient as it eliminates the need for a critic model, instead estimating the baseline from the group average of sampled outputs. 

To effectively optimize the model for complex medical scenarios, we design a composite reward function $\mathcal{R}$ consisting of three primary components: structural integrity, spatial grounding, and adaptive segmentation.

\paragraph{Format Reward ($r_{fmt}$)}
This component enforces the structural correctness of the reasoning process. The reward is 1 if the generated sequence strictly follows the required structure:
\begin{itemize}
    \item A reasoning chain within \texttt{<think>} tags;
    \item A final conclusion within \texttt{<answer>} tags;
    \item Explicit use of the \texttt{<seg>} token to trigger the segmentation head.
\end{itemize}
Non-compliant outputs receive a zero reward, ensuring the model adheres to the clinical reasoning protocol.

\paragraph{Bipartite Matching Reward ($r_{bbox}$)}
To handle \textbf{multi-lesion detection}, we formulate localization as a set prediction problem. We first employ the \textbf{Hungarian algorithm} to find the optimal bipartite matching between predicted boxes $\hat{B}$ and ground-truth boxes $B$ by minimizing the global IoU cost. 

The reward integrates spatial accuracy ($mIoU_{matched}$) with the $F_1$ score to balance precision and recall:
\begin{equation} 
r_{bbox} = mIoU_{matched} \times F_1, \quad F_1 = \frac{2 \cdot P \cdot R}{P + R}
\end{equation} 
This formulation penalizes missed lesions and false positives, helping align the model’s behavior with practical clinical expectations.

\paragraph{Dual-Granularity Mask Reward ($r_{mask}$)}
To mitigate \textbf{reward sparsity} (where $Dice \approx 0$), we propose a mechanism that shifts the focus from approximate localization to fine pixel level feedback. This ensures the policy receives meaningful signals throughout the transition from discovery to precision.

We first define the \textbf{Generalized IoU (GIoU)} for the derived bounding boxes ($B_{\hat{M}}, B_{M}$) of the mask ($\hat{M},M$) to quantify the positional discrepancy between the prediction and ground-truth:
\begin{equation} 
GIoU = IoU_{box} - \frac{|C \setminus (B_{\hat{M}} \cup B_{M})|}{|C|},
\end{equation} 
where $C$ denotes the smallest enclosing box covering both $B_{\hat{M}}$ and $B_M$. The adaptive reward is then formulated as:
\begin{equation} 
r_{mask} =
\begin{cases} 
  GIoU(B_{\hat{M}}, B_{M}), & \text{if } Dice < \tau \\ 
  1 + \lambda \cdot Dice(\hat{M},M), & \text{otherwise} 
\end{cases} 
\end{equation} 

We set the threshold $\tau=0.05$ and scaling factor $\lambda=2$. This mechanism ensures that GIoU provides an informative gradient during early training stages or for difficult, non-overlapping samples, while prioritizing pixel-perfect precision once approximate localization is achieved.

\subsubsection{Training Objective}
The second stage of our framework optimizes the model through a joint objective that aligns reasoning with segmentation, as is shown in Fig.~\ref{fig:framework}. For each input pair $(I, Q)$, the policy $\pi_{\theta}$ samples a group of $G$ responses $\{C_i\}_{i=1}^{G}$, where each response $C_i$ comprises the generated MLLM output $\hat{O}_i$ and the corresponding predicted mask $\hat{M}_i$. 

To refine the policy model without a value network, we utilize the GRPO framework. The advantage $A_i$ is computed by normalizing rewards within each group:$A_i = \frac{\mathcal{R}_i - \bar{\mathcal{R}}}{\sigma_{\mathcal{R}}},$ where $\bar{\mathcal{R}}$ and $\sigma_{\mathcal{R}}$ denote the mean and standard deviation of rewards in the group, respectively. The GRPO loss is then defined as:
\begin{equation}
\begin{split}
\mathcal{L}_{\text{GRPO}} = {} & -\frac{1}{G} \sum_{i=1}^G \Bigl[ \min \bigl( \rho_i A_i, \\
& \text{clip}(\rho_i, 1-\epsilon, 1+\epsilon) A_i \bigr) - \beta \mathbb{D}_{\text{KL}} \Bigr],
\end{split}
\end{equation}
where $\rho_i = \frac{\pi_{\theta}(\hat{O}_i | I, Q)}{\pi_{\theta_{\text{old}}}(\hat{O}_i | I, Q)}$ is the probability ratio between the current and old policies.
Here, $\epsilon$ and $\beta$ are hyperparameters for clipping and KL divergence regularization.

To maintain pixel-level accuracy during reasoning refinement, we supervise mask predictions with ground-truth $M$. The total training objective $\mathcal{L}_{\text{total}}$ is a weighted combination of the policy gradient and the group-averaged segmentation loss:
\begin{equation}
\mathcal{L}_{\text{total}} = \lambda_3 \mathcal{L}_{\text{GRPO}} + \frac{\lambda_4}{G} \sum_{i=1}^G \mathcal{L}_{\text{DiceCE}}(\hat{M}_i, M),
\end{equation}
where $\mathcal{L}_{\text{DiceCE}}$ denotes the Dice and Cross-Entropy loss, and $ \{\lambda_i\}_{i=3}^4 $ are balancing coefficients. This joint optimization empowers the model with stronger lesion exploration capability, enhanced visual reasoning, and consistent segmentation accuracy.

\begin{table*}[t]
    \centering
    \caption{\textbf{Quantitative comparison on complex lesion segmentation.} We report mDice, mIoU, and Failure Rate (Fail). Detailed Dice scores for 9 anatomical regions are provided to analyze local grounding capabilities. All values are in percentage (\%). Best results are \textbf{bolded}, second best are \underline{underlined}. $^{\dagger}$ denotes the models fine-tuned with our ComLesion-14K.}
    \label{tab:comparison}
    {
        \setlength{\tabcolsep}{3pt} 
        \begin{tabular}{l ccc c ccccccccc}
            \toprule
            \multirow{2}{*}{\textbf{Method}} & \multicolumn{3}{c}{\textbf{Overall Performance (\%)}} & & \multicolumn{9}{c}{\textbf{Anatomy-specific Dice (\%)}} \\
            \cmidrule{2-4} \cmidrule{6-14}
             & \textbf{mDice}~$\uparrow$ & \textbf{mIoU}~$\uparrow$ & \textbf{Fail}~$\downarrow$ & & \textbf{Head} & \textbf{Abd.} & \textbf{Lung} & \textbf{Eye} & \textbf{Neck} & \textbf{Skin} & \textbf{Foot} & \textbf{Breast} & \textbf{Heart} \\
             
            \midrule
            \multicolumn{14}{l}{\textbf{\textit{General MLLMs}}} \\
            Qwen2.5-VL-72B & 11.04 & 7.78 & 37.61 & & 6.68 & 5.11 & 3.30 & 5.23 & 31.36 & 54.09 & 22.89 & 52.86 & 1.21 \\
            InternVL3-8B & 4.18 & 2.72 & 73.90 & & 2.36 & 0.50 & 0.79 & 0.96 & 11.34 & 31.44 & 5.55 & 14.07 & 0.86 \\
            Qwen2.5-VL-7B & 8.30 & 5.56 & 42.76 & & 3.62 & 4.47 & 2.62 & 2.89 & 23.79 & 48.12 & 5.18 & 41.43 & 3.06 \\
            
            \midrule
            \multicolumn{14}{l}{\textbf{\textit{Medical-Specific MLLMs}}} \\
            HuatuoGPT-7B & 4.73 & 3.04 & 43.72 & & 3.58 & 1.88 & 2.02 & 3.20 & 13.49 & 20.13 & 0.10 & 16.11 & 0.62 \\
            Chiron-o1-8B & 3.35 & 2.19 & 66.61 & & 2.35 & 0.81 & 1.50 & 1.86 & 9.60 & 14.74 & 1.80 & 28.01 & 0.84 \\
            Lingshu-7B & 7.37 & 4.73 & 58.41 & & 3.34 & 4.62 & 2.60 & 3.27 & 22.56 & 37.76 & 6.01 & 26.61 & 0.88 \\
            MedGemma-4B & 6.03 & 3.85 & 49.11 & & 3.10 & 1.95 & 2.39 & 2.30 & 19.09 & 34.15 & 4.08 & 25.99 & 1.98 \\
            
            \midrule
            \multicolumn{14}{l}{\textbf{\textit{Grounding-Specific MLLMs}}} \\
            Seg-Zero-7B & 9.08 & 6.23 & \underline{19.84} & & 4.72 & 2.49 & 3.72 & 3.15 & 25.87 & 54.57 & 7.53 & 50.33 & 2.92 \\
            VLMR1-REC-3B & 4.76 & 2.99 & 24.98 & & 4.45 & 1.14 & 2.76 & 3.50 & 11.74 & 15.65 & 1.69 & 32.77 & 1.27 \\
            MedPLIB-7B & 12.90 & 8.77 & 42.40 & & 12.51 & 6.00 & 4.12 & 4.82 & 22.11 & 44.45 & \textbf{48.34} & 44.85 & 3.20 \\
            MediSee-7B& 11.07 & 7.49 & 55.97 & & \underline{21.18} & 4.01 & 4.00 & 1.20 & 5.16 & 22.98 & 9.90 & 5.58 & \underline{9.84} \\
            Medical SAM3 & 16.58 & 12.25 & 53.66 & & 20.5 & 3.64 & \underline{19.75} & 3.93 & 23.07 & 53.98 & 27.31 & \underline{58.89} & 2.02 \\
            LISA-3B $^{\dagger}$& \underline{22.17} & \underline{16.49} & 44.28 & & 16.39 & \underline{22.71} & 19.40 & 5.64 & 35.51 & \underline{66.20} & 17.89 & 48.21 & 2.39 \\
            Seg-Zero-3B $^{\dagger}$ & 17.47 & 12.28 & 52.57 & & 13.41 & 13.94 & 14.37 & \textbf{8.72} & \underline{40.64} & 53.54 & \underline{39.51} & 21.78 & \textbf{12.21} \\
            \textbf{CORE-Seg (Ours)} & \textbf{37.06} & \textbf{27.79} & \textbf{18.42} & & \textbf{31.40} & \textbf{39.56} & \textbf{41.40} & \underline{8.28} & \textbf{47.25} & \textbf{74.85} & 37.53 & \textbf{70.99} & 8.14 \\
            \bottomrule
        \end{tabular}
    }
\end{table*}

\begin{figure*}[!t]
\centerline{\includegraphics[width=\linewidth]{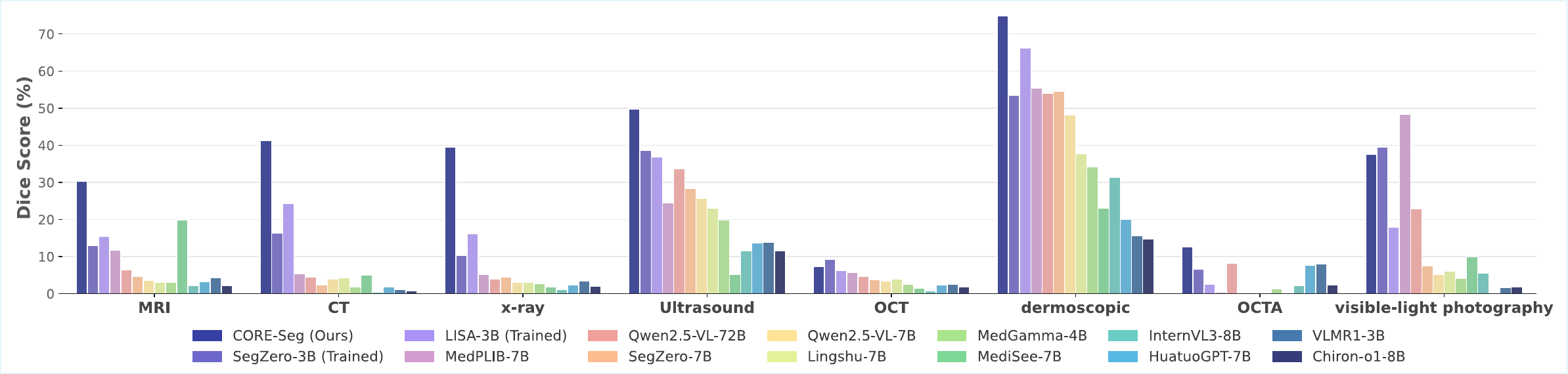}}
\caption{\textbf{Comparison of mDice scores across 8 medical modalities.}  CORE-Seg demonstrates superior generalization capabilities, achieving state-of-the-art performance particularly among MLLM-based methods in MRI, CT, X-Ray and Ultrasound.}
\label{fig:modality_bar}
\end{figure*}

\section{Experiments}
\subsection{Data and Experimental Setup}
\subsubsection{Datasets}
\label{ood}
We evaluate our method on the ComLesion-14K dataset, which is described in detail in Sec.~\ref{sec:dataset} and Table~\ref{tab:composition}. To avoid data leakage, duplicate and near-duplicate images were removed during preprocessing. For 3D volumetric data, we sampled slices at an interval of 2 to reduce the high visual similarity between adjacent slices from the same patient. For the MLLM input, all images are resized so that the longer edge is at most 512 pixels while preserving the aspect ratio. For the SAM input, images are resized to 512$\times$512. We randomly sample 2,421 image-text-mask triplets for the test set, and use the remaining data for training. The training and test sets were split at the patient level to ensure strict independence.

To further assess the generalization of our two-stage training strategy, we categorize \textbf{Out-of-Domain (OOD) benchmarks} into three levels by the severity of distribution shift:
(1) TN-SCUI2020~\cite{kang2022thyroid} (Ultrasound thyroid nodule): For evaluating basic distribution shifts. (2) ISPY~\cite{chitalia2022expert} (MRI breast tumor): With an unseen novel lesion class. (3) CVC-ClinicDB~\cite{bernal2015wm} (Endoscopic polyp): The most challenging, with domain gaps in both imaging modality and semantic class.

\begin{table*}[t]
\centering
\caption{\textbf{Comparison with specialist segmentation models.} In-domain (ID) results are on ComLesion-14K; OOD results are mDice (\%) on TN-SCUI, CVC-ClinicDB, and ISPY. All specialist models are trained on the same patient-level training split of ComLesion-14K. ``--'' indicates that the multi-class model cannot be evaluated on OOD data due to label-space mismatch.}
\label{tab:specialist}
{%
\setlength{\tabcolsep}{3.5pt}
\begin{tabular}{l|ccc|ccc}
\toprule
\multirow{2}{*}{\textbf{Method}} & \multicolumn{3}{c|}{\textbf{ID Performance}} & \multicolumn{3}{c}{\textbf{OOD Robustness (mDice $\uparrow$)}} \\
\cmidrule(lr){2-4} \cmidrule(l){5-7}
 & \textbf{mDice $\uparrow$} & \textbf{mIoU $\uparrow$} & \textbf{Fail $\downarrow$} & \textbf{TN-SCUI} & \textbf{CVC-ClinicDB} & \textbf{ISPY} \\ \midrule
nnU-Net (Binary) & 22.05 & 16.92 & 59.27 & 63.06 & 5.75 & 0.00 \\
nnU-Net (Multi-class) & 40.89 & 31.65 & 25.36 & 58.21 & -- & -- \\
LViT (Referring) & 41.25 & 33.97 & 24.32 & 22.87 & 1.10 & 1.08 \\
\textbf{CORE-Seg (Reasoning)} & 37.06 & 27.79 & \textbf{18.42} & \textbf{74.82} & \textbf{48.20} & \textbf{38.71} \\
\bottomrule
\end{tabular}%
}
\end{table*}

\subsubsection{Implementation Details}
To implement the difficulty-aware filtering strategy in dataset generation, we train three independent U-Net models with identical architectures but different random initialization seeds for each lesion class to mitigate selection bias. These U-Net models adopt ResNet-18 as the encoder pre-trained on the ImageNet dataset, with an input resolution of \(512 \times 512\) and 3 input channels for single-class lesion segmentation output. We use the Adam optimizer with a learning rate of \(10^{-4}\) and a batch size of 16, and Binary Dice loss is taken as the loss function.

Our framework utilizes Qwen2.5-VL-3B~\cite{bai2025qwen2} as the reasoning backbone and MedSAM2 for segmentation, implemented on 2 NVIDIA A6000 GPUs using AdamW optimizer with DeepSpeed acceleration. We employ LoRA ($r=64, \alpha=128, \text{dropout}=0.05$) for the MLLM, while fully fine-tuning the Semantic-Guided Prompt Adapter and Mask Decoder. The training proceeds in two stages within the same unified framework, and the segmentation loss remains consistent across both stages. In the first stage, we train for 1 epoch with a batch size of 32 and a learning rate of $1\text{e-}4$. In the second stage, we fine-tune for 2 epochs with a reduced learning rate of $5\text{e-}6$ and a batch size of 16. The sampling number is set to 4. The loss weights $\{\lambda_i\}_{i=1}^4$ are empirically set to 1.0.

\subsubsection{Evaluation Metrics}
To comprehensively assess segmentation quality and system robustness, we employ the \textbf{Dice Similarity Coefficient (Dice)} and \textbf{Intersection over Union (IoU)} to measure the spatial overlap between predicted masks and ground truth. However, given the high complexity of lesion segmentation tasks, standard MLLMs are prone to severe hallucinations or instruction-following failures. To explicitly quantify such instability, we introduce the \textbf{Failure Rate} as a critical reliability metric. A prediction is classified as a failure if: (1) the resulting Dice score is 0 (indicating a complete loss of visual grounding), or (2) the MLLM generates an output with an invalid format.

\subsection{Complex Lesion Reasoning Segmentation Results}
\subsubsection{Baselines}
To comprehensively validate CORE-Seg, we conduct extensive comparisons with a wide array of state-of-the-art models across three categories: (1) \textbf{General MLLMs}: Qwen2.5-VL-72B, Qwen2.5-VL-7B, and InternVL3-8B~\cite{zhu2025internvl3}; (2) \textbf{Medical-Specific MLLMs}: HuatuoGPT-7B~\cite{chen2024towards}, Chiron-o1-8B~\cite{sun2025chiron}, Lingshu-7B~\cite{xu2025lingshu}, and MedGemma-4B~\cite{sellergren2025medgemma}; and (3) \textbf{Grounding-Specific MLLMs}: Seg-Zero-7B, VLMR1-REC-3B~\cite{shen2025vlm}, MedPLIB-7B, MediSee-7B, and Medical SAM3~\cite{jiang2026medical}. Specifically, the models above are evaluated in a zero-shot manner to assess their intrinsic medical adaptability using their official pre-trained weights. Furthermore, for a strictly fair comparison, we fine-tuned LISA-3B and Seg-Zero-3B aligning them with the same Qwen2.5-VL-3B backbone and MedSAM2 to validate our framework advantages.

Beyond MLLM-based methods, we also compare CORE-Seg with representative \textbf{specialist segmentation models}, including nnU-Net~\cite{isensee2021nnu} (binary and multi-class) and LViT~\cite{li2023lvit}, trained on the same patient-level split of ComLesion-14K. This additional comparison clarifies the performance boundary between reasoning-based MLLMs and single-task segmentation paradigms.

\begin{table*}[!h]
\centering
\caption{\textbf{Ablation study on progressive framework.} We report mDice, mIoU, and Failure Rate (Fail) on ID data. For OOD robustness, we report mDice and Fail across three challenging datasets: CVC-ClinicDB, TN-SCUI, and ISPY.}
\label{tab:ablation}
{%
\setlength{\tabcolsep}{3.5pt} 
\begin{tabular}{ccc|ccc|ccc}
\toprule
\multicolumn{3}{c|}{\textbf{Components}} & \multicolumn{3}{c|}{\textbf{ID Performance}} & \multicolumn{3}{c}{\textbf{OOD Robustness (mDice$\uparrow$/Fail$\downarrow$)}} \\
\cmidrule(r){1-3} \cmidrule(lr){4-6} \cmidrule(l){7-9}
\textbf{SG-A} & \textbf{S1} & \textbf{S2} & \textbf{mDice $\uparrow$} & \textbf{mIoU $\uparrow$} & \textbf{Fail $\downarrow$} & \textbf{CVC-ClinicDB} & \textbf{TN-SCUI} & \textbf{ISPY} \\ \midrule
- & - & - & 4.86 & 3.00 & 58.5 & 30.35/27.6 & 15.01/37.8 & 0.98/65.2 \\
- & \checkmark & - & 23.60 & 16.57 & 45.2 & 37.67/26.2 & 50.26/14.4 & 32.26/21.6 \\
\checkmark & \checkmark & - & 30.37 & 22.20 & 26.02 & 34.16/17.20 & 64.16/\textbf{1.60} & 29.31/15.40 \\
\checkmark & \checkmark & \checkmark & \textbf{37.06} & \textbf{27.79} & \textbf{18.42} & \textbf{48.20/15.00} & \textbf{74.82}/2.20 & \textbf{38.71/13.00} \\ \bottomrule
\end{tabular}%
}
\end{table*}

\begin{table*}[!h]
    \centering
    \caption{\textbf{Ablation study of adaptive reward mechanism in CORE-Seg.}
    We analyze the impact of training-time reasoning guidance, reward types, Dice loss, and explicitly quantify the effect of reasoning quality (Random vs. Description-only vs. Full CoT).}
    \label{tab:ablation_reward}
    {%
    \begin{tabular}{ccccc|ccc}
        \toprule
        \textbf{CoT} & \textbf{$r_{fmt}$} & \textbf{$r_{bbox}$} & \textbf{$r_{mask}$} & \textbf{$\mathcal{L}_{\text{DiceCE}}$} & \textbf{mDice $\uparrow$} & \textbf{mIoU $\uparrow$} & \textbf{Fail $\downarrow$} \\
        \midrule
        -- & -- & -- & -- & -- & 31.24 & 23.15 & 20.08 \\   
        \checkmark & -- & -- & -- & -- & 30.37 & 22.20 & 26.02 \\ 
        \midrule
        \checkmark & \checkmark & \checkmark & -- & -- & 26.87 & 18.91 & 30.08 \\
        \checkmark & \checkmark &  -- & \checkmark & -- & 31.83 & 23.45 & 24.63 \\ 
        \checkmark & \checkmark & \checkmark & \checkmark & -- & 32.41 & 24.06 & 24.32 \\ 
        \midrule
        Rand. & \checkmark & \checkmark & \checkmark & \checkmark & 33.85 & 24.91 & 23.64 \\ 
        Desc. & \checkmark & \checkmark & \checkmark & \checkmark & 35.78 & 26.54 & 19.85 \\ 
        \checkmark & \checkmark & \checkmark & \checkmark & \checkmark & \textbf{37.06} & \textbf{27.79} & \textbf{18.42} \\
        \bottomrule
    \end{tabular}
    }
\end{table*}

\subsubsection{Overall Performance}
In Table~\ref{tab:comparison}, CORE-Seg establishes a new state-of-the-art (SOTA) among MLLM-based methods for complex lesion segmentation with \textbf{37.06\% mDice} and \textbf{27.79\% mIoU}, outperforming the best MLLM baseline LISA-3B by +14.89\%. While General and Medical MLLMs suffer high failure rates due to weak grounding alignment, CORE-Seg maintains the lowest failure rate at \textbf{18.42\%} among all compared MLLM-based models.

Notably, existing grounding MLLMs (MedPLIB, MediSee) show limited performance ($\sim$12\% mDice) due to scarce lesion samples in their organ-centric training data, verifying the necessity of our lesion-targeted model. Moreover, comparisons between fine-tuned models highlight our architectural advantages: SFT-based LISA has a high failure rate (44.28\%) and is prone to overfitting; cascaded Seg-Zero matches our low failure rate but suffers low accuracy from multi-stage error accumulation. Ultimately, CORE-Seg proves an end-to-end RL-driven paradigm effectively balances precise localization and generalization.

We further compare CORE-Seg with specialist segmentation models, including nnU-Net (binary and multi-class) and LViT, trained on the same ComLesion-14K, using their official open‑source repository default hyper‑parameters. As shown in Table~\ref{tab:specialist}, nnU-Net (multi-class) and LViT achieve higher in-domain mDice than CORE-Seg, because they are single-task models optimized purely for pixel-level segmentation within a fixed label space. However, they degrade under domain shift, whereas CORE-Seg maintains stronger OOD performance. This confirms that CORE-Seg is not intended to surpass specialist models in closed-set in-domain overlap, but offers superior generalization and interpretability for complex, unseen clinical scenarios.

\subsubsection{Anatomical and Modal Analysis} 
As shown in Table~\ref{tab:comparison} and Fig.~\ref{fig:modality_bar}, CORE-Seg exhibits strong cross-distribution robustness in complex lesion segmentation: it outperforms baselines in 6 of 9 anatomical categories, adapts well to grayscale radiology modalities (MRI/CT) where general MLLMs fail, and maintains robust segmentation in high-noise modalities (Ultrasound/OCT), validating the universal feature enhancement of its reasoning mechanism. 

\subsubsection{Parameter Efficiency Analysis} 
Crucially, CORE-Seg achieves SOTA performance among reasoning-based MLLM frameworks with exceptional efficiency. Built upon a compact 3B backbone optimized via LoRA, our model operates with minimal trainable parameters and computational overhead. Despite being ${24\times}$ smaller than the massive Qwen2.5-VL-72B, CORE-Seg surpasses it by 26.02\% in mDice. This economical training strategy proves that explicit reasoning alignment outweighs sheer parameter scaling, enabling rapid deployment for complex medical segmentation even in resource-constrained clinical environments.

\subsection{Ablation Study}

\subsubsection{Ablation Study on Progressive Framework}
Table~\ref{tab:ablation} quantifies the efficacy of our progressive design by dissecting module and training stage contributions from two perspectives.

We first evaluate the \textbf{Semantic-Guided Prompt Adapter (SG-A)} on the in domain dataset. In SFT-Stage (S1), replacing SG-A with a simple \textit{Linear layer} drops mDice from 30.37\% to 23.60\%, demonstrating that SG-A is essential for aligning semantic priors whereas naive projections are insufficient. However, relying solely on SFT tends to overfit the training distribution, limiting performance on complex cases. Subsequently, incorporating RL-Stage (S2) boosts mDice to 37.06\% and minimizes the failure rate to 18.42\%, confirming that S2 significantly enhances generalization, effectively refining details.

Then we test \textbf{OOD robustness} on TN-SCUI, ISPY and CVC-ClinicDB (as Sec.~\ref{ood} describes). S1 has basic capabilities but overfits the source domain, restricting generalization. As shown in Table~\ref{tab:ablation}, despite the increasing severity of domain shifts across these datasets, S2 consistently outperforms the SFT baseline by approximately 10\% mDice (e.g., +10.66\% on TN-SCUI, +9.4\% on ISPY, and +14.04\% on CVC-ClinicDB). This improvement confirms that the exploration mechanism in RL is critical for overcoming SFT-induced overfitting, enabling the model to learn robust, transferable policies.

\begin{table*}[!h]
    \centering
    \caption{\textbf{Ablation study on the semantic probe selection.} We compare using the specific \texttt{<seg>} token (Ours) versus the hidden states of the whole answer across training stages. }
    {%
    \label{tab:ablation_probe}
    \begin{tabular}{l|c|ccc}
        \toprule
        \textbf{Semantic Probe} & \textbf{Training Stage} & \textbf{mDice $\uparrow$} & \textbf{mIoU $\uparrow$} & \textbf{Fail $\downarrow$} \\
        \midrule
        \multirow{2}{*}{\textit{Whole Answer}} 
            & S1 & 32.25 & 23.90 & 24.82 \\
            & S1 + S2 & 33.01 & 24.54 & 24.32 \\
        \midrule
        \multirow{2}{*}{\textbf{\texttt{<seg>} Token}} 
            & S1 & 30.37 & 22.20 & 26.02 \\
            & S1 + S2 & \textbf{37.06} & \textbf{27.79} & \textbf{18.42} \\
        \bottomrule
    \end{tabular}
    }
\end{table*}

\subsubsection{Ablation Study on Adaptive Reward Mechanism}

This ablation study verifies the adaptive reward mechanism of CORE-Seg from three perspectives: optimization strategy, reward configuration, and reasoning quality.

We first investigate the synergy between reasoning and the \textbf{optimization strategy}. Under the standard SFT setting, simply appending a CoT slightly decreases the mDice by 0.87\% compared to the baseline. This suggests that standard supervised learning struggles to fully exploit the intermediate reasoning steps. However, when optimized with GRPO, the model's reasoning capability is effectively unlocked, improving the mDice to 37.06\%. This demonstrates that explicit reasoning requires reward-based optimization to genuinely benefit the final segmentation.

We further analyze the \textbf{specific contributions of the rewards} proposed for Stage 2. Relying exclusively on the $r_{bbox}$ reduces the mDice to 26.87\%, as the objective shifts from fine segmentation to coarse localization. Conversely, using the $r_{mask}$ alone achieves 31.83\%, underscoring the importance of dense feedback at the mask level. The combination of both rewards yields 32.41\%, suggesting that $r_{bbox}$ provides spatial constraints that complement the prediction of the \texttt{seg} token. Ultimately, integrating $\mathcal{L}_{\text{DiceCE}}$ with RL rewards leads to a peak performance of 37.06\% mDice. This synergy is vital, while RL exploration identifies optimal policies, $\mathcal{L}_{\text{DiceCE}}$ offers stable gradients at the pixel level to mitigate variance and refine boundaries.

To isolate \textbf{the impact of reasoning}, we modified the CoT content (Table~\ref{tab:ablation_reward}). Substituting correct rationales with random samples (\textbf{Rand.}) drops the mDice to 33.85\% and increases the failure rate to 23.64\%, confirming that the model relies on valid semantic guidance rather than sequence length. Furthermore, using purely descriptive text (\textbf{Desc.}) without sequential deduction results in an mDice of 35.78\%. While accurate descriptions provide a strong baseline, they trail our full reasoning model by 1.28\%. This performance gap proves that explicit logical deduction provides the essential constraints required for precise localization and mask refinement.

\subsubsection{Effect of Semantic Probe Selection}
We investigate the choice of semantic probes for the segmentation head: using a dedicated \texttt{<seg>} token versus the hidden states of the entire answer. As shown in Table~\ref{tab:ablation_probe}, in Stage 1, employing the whole answer as the semantic probe yields better initial performance, likely due to its access to richer contextual semantics.

However, during RL-stage, the \texttt{<seg>} token exhibits clear advantages, achieving a final mDice of 37.06\%, significantly outperforming the whole answer variant. This benefit arises from two factors: (1) the token appears at the end of the sequence, enabling its hidden state to aggregate global reasoning context; and (2) being generated right after localization outputs, it naturally captures localization-aware information that aligns detection and segmentation. Furthermore, under GRPO training, using the \texttt{<seg>} tends to yield higher reward variance, providing stronger discriminative guidance for policy updates. This leads to more stable learning and ultimately better segmentation performance with fewer failures.

\begin{figure*}[!t]
\centerline{\includegraphics[width=\linewidth]{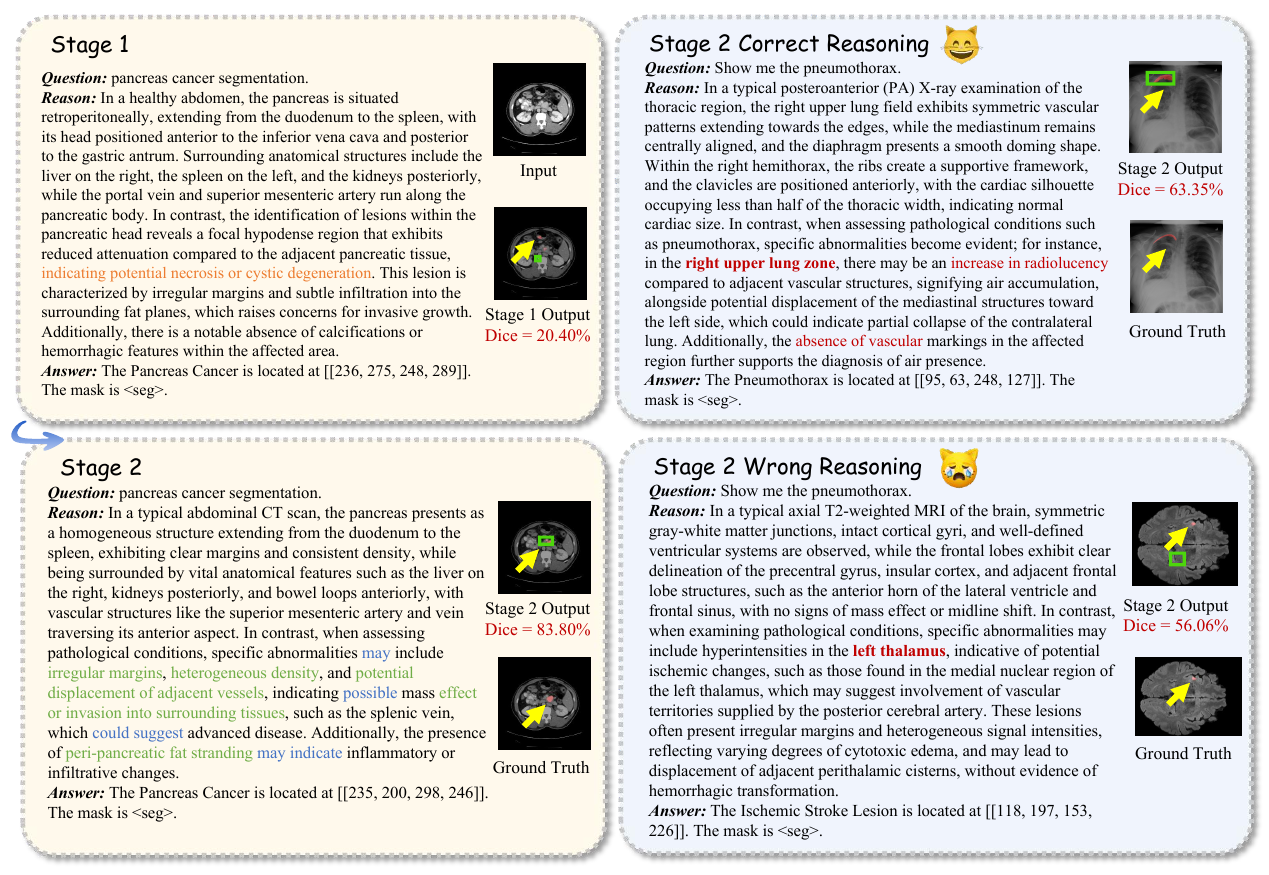}}
\caption{\textbf{Qualitative visualization and reasoning analysis of our two-stage framework.} 
The left part illustrates that Stage 2 refines the outputs of Stage 1 via image-grounded reasoning, yielding higher-quality masks and stronger robustness, especially for small lesions and cases with inaccurate bounding box localization. 
The right part presents typical correct and erroneous reasoning pipelines, demonstrating that both bounding box localization and mask prediction are explicitly influenced by visual reasoning descriptions.}
\label{fig:visualize_cot}
\end{figure*}

\begin{table}[h]
\centering
\caption{\textbf{Comparison of different SAM backbones.} We evaluate the impact of the SAM version in Stage1.}
\label{tab:sam_version}
{%
\begin{tabular}{l|ccc}
\toprule
\textbf{Segmentation Model} & \textbf{mDice $\uparrow$} & \textbf{mIoU $\uparrow$} & \textbf{Fail $\downarrow$} \\ \midrule
SAMed2D & 19.58 & 13.44 & 45.28 \\
SAM 2 & 23.60 & 16.57 & 39.87 \\
\textbf{MedSAM2 (Ours)} & \textbf{30.37} & \textbf{22.20} & \textbf{26.02} \\ \bottomrule
\end{tabular}%
}
\end{table}

\subsubsection{Backbone Scalability}
We compare different SAM backbones at Stage 1 to isolate the impact of the segmentation module. Table~\ref{tab:sam_version} shows that MedSAM2 outperforms SAMed2D and SAM 2~\cite{ravi2024sam}, exhibiting superior robustness on our medical dataset. Thus, we adopt MedSAM2 as the default segmentation model to achieve optimal semantic prompting compatibility and the lowest failure rate.

\subsection{Qualitative Analysis}
\subsubsection{Progressive Improvements in Reasoning and Masks}

Our case analysis in Fig.~\ref{fig:visualize_cot} (left) reveals a clear transition from template-like reasoning in Stage 1 to deliberate, spatially grounded reasoning in Stage 2, which directly correlates with improved segmentation quality. While Stage 1 reasoning frequently yields \emph{hallucinated} or overconfident visual descriptions that lead to coarse, misplaced masks, Stage 2 produces cautious, image-grounded explanations. By incorporating uncertainty-aware expressions (e.g., \emph{possible}, \emph{may indicate}) and spatially verifiable anatomical cues, Stage 2 establishes tight consistency between the model's textual logic and the final segmentation outputs.

To analyze the role of spatial reasoning, we visualize two representative cases in Fig.~\ref{fig:visualize_cot} (right). In the pneumothorax case, the Stage 2 rationale correctly identifies abnormalities in the \emph{right upper lung zone}, consistent with the predicted mask under the standard chest X-ray mirror view. Conversely, in the brain lesion case, despite an incorrect textual and bounding-box localization pointing to the left thalamus, the model still successfully segments the actual lesion in the right frontal region. Together, these cases demonstrate that CORE-Seg does not simply propagate errors in a cascaded manner. Instead, it encourages mutual consistency between spatial reasoning and mask prediction, while leveraging visual features to robustly correct textual or box-level localization inaccuracies.

For clinical validation, we randomly selected 200 samples for \textbf{expert review} by senior clinicians. The assessment revealed that in 169 cases (84.5\%), the generated reasoning was generally aligned with the regional characteristics of the corresponding segmentations or bounding boxes without significant discrepancies. This evaluation suggests that the reasoning process exerts a positive influence on the segmentation and localization regions, playing a supportive role in guiding the model toward more consistent boundaries.

\begin{figure}[!t]
\centerline{\includegraphics[width=\columnwidth]{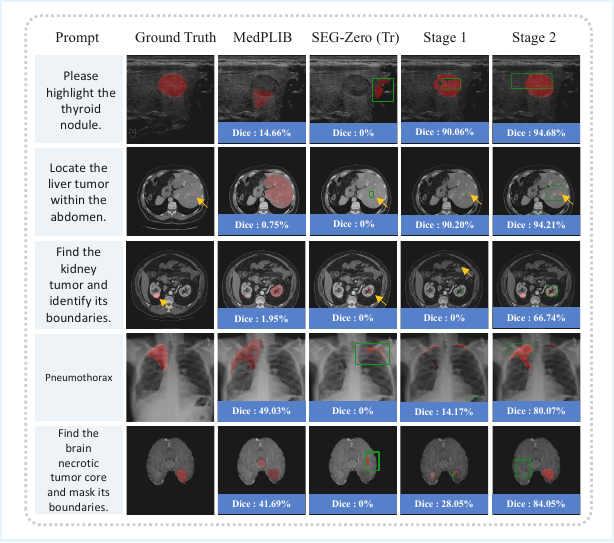}}
\caption{\textbf{Visualization of our method with baselines.}}
\label{fig:visualize_seg}
\end{figure}

\subsubsection{Comparison with Baselines}
Compared to recent strong baselines, our method shows more robust performance across challenging cases (see Fig.~\ref{fig:visualize_seg}). MedPLIB misses small lesions and often degenerates to organ-level segmentation, while Seg-Zero (Trained) relies heavily on bounding-box localization with error accumulation from inaccurate boxes.
In contrast, our approach effectively decouples localization and segmentation. Even with imprecise boxes, it recovers correct regions via reasoning-guided refinement, delivering superior masks and higher Dice scores across modalities and targets.

\begin{figure}[!ht]
\centering
\includegraphics[width=0.96\columnwidth]{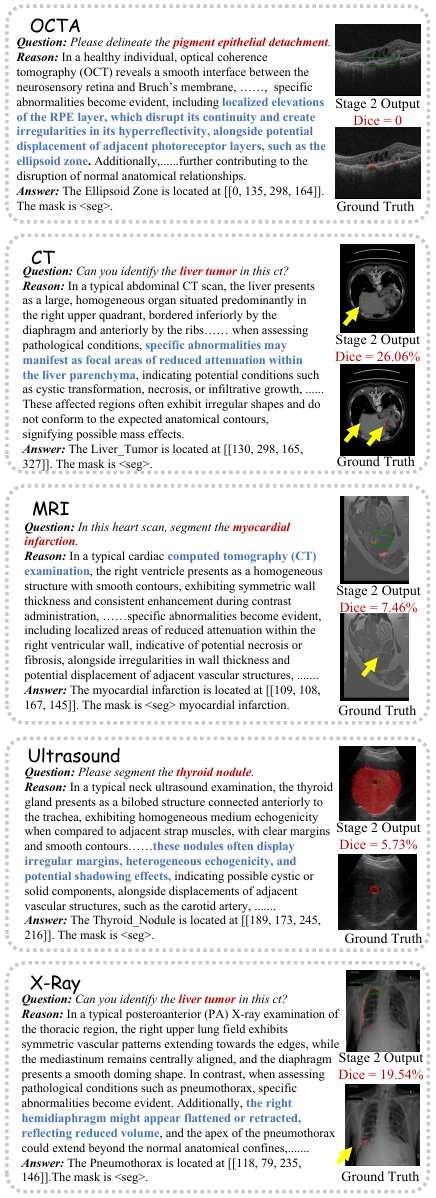}
\caption{\textbf{Failure case study.}}
\label{fig:failure_cases}
\end{figure}

\subsubsection{Failure Case Study}
To analyze the limitations of our method, we visualize 5 representative failure cases across various imaging modalities in Fig. \ref{fig:failure_cases}. These failures primarily include under-segmentation, over-segmentation, and subregion misclassification. Specifically, the model occasionally misses lesion parts, includes surrounding healthy tissue, or confuses distinct lesion types. These errors are largely driven by the inherent complexity of the selected samples.

Furthermore, reasoning errors occur in the Chain-of-Thought process. For example, the model may misidentify MRI scans as CT imaging in the cardiac EMIDEC dataset. This misperception is likely due to the limited training data available for these specific scenarios.

\section{Conclusion}
In this paper, we explore the transition from perceptual segmentation to reasoning-driven analysis for complex medical lesions. To this end, we construct ComLesion-14K, the first large-scale benchmark specifically designed to capture challenging clinical scenarios that require explicit medical reasoning, filling a critical gap in existing datasets. To address the associated challenge, we developed CORE-Seg, an end-to-end framework with a Semantic-Guided Prompt Adapter that links medical logic to segmentation via latent interaction. This method adopts a progressive SFT-to-GRPO training strategy equipped with an adaptive dual-granularity reward, which alleviates reward sparsity while enhancing reasoning interpretability and segmentation accuracy. Experiments show that it achieves SOTA performance among reasoning-based MLLM frameworks and reduces the failure rate, suggesting its potential for clinical use.

Despite the encouraging results, several limitations remain. The current framework mainly targets 2D images and does not support 3D volumetric medical data, limiting its ability to exploit inter-slice context. In addition, inference can be slow due to explicit reasoning, which increases computational cost. Future work will extend the framework to volumetric segmentation and improve efficiency while maintaining accuracy and clinical reliability.

\subsection*{Availability of Data and Materials}
The data that support the findings of this study can be downloaded from \href{https://xyx1024.github.io/CORE-Seg.github.io/}{https://xyx1024.github.io/CORE-Seg.github.io}.

\subsection*{Author Contributions}

Yuxin Xie contributed to methodological development, dataset construction, coding, experiments, data analysis and interpretation, as well as the writing of the original manuscript. Yuming Chen and Yishan Yang both participated in dataset construction, coding, experiments, and data analysis and interpretation. Yi Zhou was responsible for conceptualization, funding acquisition, methodological development, data analysis and interpretation, original manuscript writing, and manuscript review and editing. Tao Zhou and Huazhu Fu both contributed to data analysis and interpretation, as well as manuscript review and editing. Zhen Zhao and Jiacheng Liu both performed pathological validation and contributed to data analysis and interpretation.

\subsection*{Acknowledgements}
This work was supported by the National Natural Science Foundation of China under Grant 62476054 and Grant 62576153.

\subsection*{Declaration of Competing Interest}

The authors have no competing interests to declare that are relevant to the
content of this article.

\bibliographystyle{CVMbib}
\bibliography{refs}

\end{document}